\documentclass[lettersize,journal]{IEEEtran}
\usepackage{amsmath,amsfonts}
\usepackage{algorithmic}
\usepackage{algorithm}
\usepackage{array}
\usepackage[caption=false,font=normalsize,labelfont=sf,textfont=sf]{subfig}
\usepackage{textcomp}
\usepackage{stfloats}
\usepackage{url}
\usepackage{verbatim}
\usepackage{graphicx}
\usepackage{cite}
\usepackage{multirow}
\usepackage{color,soul}
\begin{document}

\title{Integrated Multiscale Domain Adaptive YOLO}

\author{Mazin Hnewa, \IEEEmembership{Student Member, IEEE} and Hayder Radha, \IEEEmembership{Fellow, IEEE}
\thanks{This work has been supported by the Ford Motor Company under the Ford-MSU Alliance Program.}
\thanks{Mazin  Hnewa and  Hayder Radha are with Department of Electrical and Computer Engineering,  Michigan State University, East Lansing, MI 48824 USA. (e-mail: (mazin, radha)@msu.edu).}
}



\maketitle

\begin{abstract}
The area of domain adaptation has been instrumental in addressing the domain shift problem encountered by many applications. This problem arises due to the difference between the distributions of source data used for training in comparison with target data used during realistic testing scenarios. In this paper, we introduce a novel MultiScale Domain Adaptive YOLO (MS-DAYOLO) framework that employs multiple domain adaptation paths and corresponding domain classifiers at different scales of the recently introduced YOLOv4 object detector. Building on our baseline multiscale DAYOLO framework, we introduce three novel deep learning architectures for a Domain Adaptation Network (DAN) that generates domain-invariant features. In particular, we propose a Progressive Feature Reduction (PFR), a Unified Classifier (UC), and an Integrated architecture. We train and test our proposed DAN architectures in conjunction with YOLOv4 using popular datasets. Our experiments show significant improvements in object detection performance when training YOLOv4 using the proposed MS-DAYOLO architectures and when tested on target data for autonomous driving applications. Moreover, MS-DAYOLO framework achieves an order of magnitude real-time speed improvement relative to Faster R-CNN solutions while providing comparable object detection performance.
\end{abstract}


\section{Introduction}
\begin{figure}[!t]
\begin{center}
  \includegraphics[width=\linewidth]{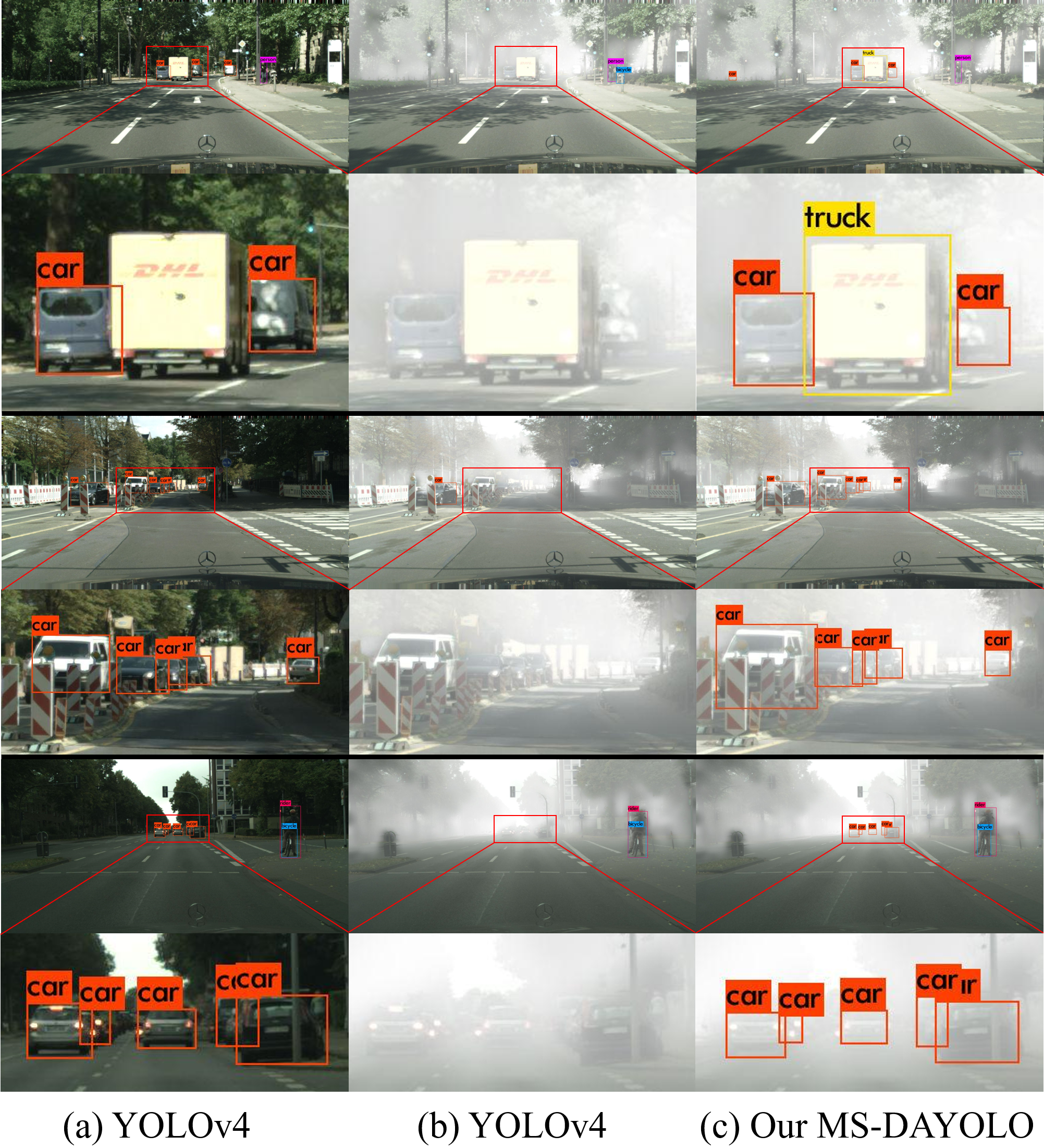}
\end{center}
  \caption{Visual detection examples using the original YOLOv4 method on: (a) clear images and (b) foggy images. (c) Our proposed MS-DAYOLO applied onto foggy images. The images are from the Cityscapes \cite{Cityscapes} and Foggy Cityscapes \cite{Foggy_cityscapes} datasets.}
\label{fig:cityscapes_examples}
\end{figure}

\IEEEPARstart{C}{onvolutional} Neural Networks (CNNs) have been achieving exceedingly improved performance for object detection in terms of classifying and localizing a variety of objects in a scene. \cite{girshick2014rich,girshick2015fast,FasterRCNN,liu2016ssd,YOLO,RetinaNet,R-FCN}. However, under a domain shift, when the testing data has a different distribution from the training data distribution, the performance of state-of-the-art object detection methods, drops noticeably and sometimes significantly. Such domain shift could occur due to capturing the data under different lighting or weather conditions, or due to viewing the same objects from different viewpoints leading to changes in object appearance and background. For example, training data used for autonomous vehicles is normally captured under favorable clear weather conditions whereas testing could take place under more challenging weather (\textit{e.g.} rain, fog). Consequently, methods fail to detect objects as shown in the examples of Figure \ref{fig:cityscapes_examples}(b). In that context, the domain under which training is done is known as the \textit{source domain} while the new domain under which testing is conducted is referred to as the \textit{target domain}.

One of the challenges that aggravates the domain shift problem is the lack of target domain data, and especially annotated data. This led to the emergence of the area of \textit {domain adaptation} \cite{duan2012domain,kulis2011you,ganin2016domain,tzeng2017adversarial,long2016unsupervised,GRL}, which has been widely studied to solve the problem of domain shift without the need to annotate data for new target domains. In general, domain adaptation solutions have relied on adversarial networks and other strategies that are designed to generate domain-invariant features. Consequently, the particular domain adaptation solution used is influenced greatly by the underlying object detection method architecture. In that context, within the area of object detection, domain adaptation has been studied rather extensively for Faster R-CNN object detection and its variants \cite{chen2018domain,zhu2019adapting,wang2019few,saito2019strong,he2019multi, sindagi2020prior}. However, other popular object detection schemes, and in particular YOLO-based architectures, have received little or no attention \cite{zhang2021domain,liu2021image}.

In this paper, we propose novel domain adaptation architectures for the YOLOv4 object detector. In particular, we introduce four new \textit{MultiScale Domain Adaptive YOLO} (MS-DAYOLO) architectures that promote multiscale domain adaptation for the feature extraction stage and progressive channel reduction strategies for domain classifiers. The proposed MS-DAYOLO framework achieves significant improvements over YOLOv4 as shown in the examples of Figure \ref{fig:cityscapes_examples}(c). MS-DAYOLO achieves an order of magnitude real-time speed improvement relative to Faster R-CNN solutions while providing comparable and in some instances superior object detection performance. In particular, the main contributions of this paper can be summarized as follows:
\begin{enumerate}
    \item \textbf{Baseline Multiscale DAYOLO$:$} We introduce a Multiscale Domain Adaptive YOLO (MS-DAYOLO) architecture that supports domain adaptation at different layers of the feature extraction stage within the YOLOv4 backbone network. The MS-DAYOLO architecture, which includes a Domain Adaptive Network (DAN) with multiscale feature inputs and multiple domain classifiers, represents our baseline domain adaptive YOLO framework. It is important to highlight that we have recently introduced the proposed baseline architecture in \cite{MS-DYOLO}. Here, we build on this prior work by: (a) introducing three new novel MS-DAYOLO architectures, (b) presenting significantly improved and new performance results relative to the baseline performance, and (c) conducting new and more extensive simulation and ablation studies.
    \item \textbf{Progressive Feature Reduction, Unified Classifier, and Integrated Multiscale DAYOLO$:$} Building on our baseline MS-DAYOLO framework, we propose three novel domain adaptation architectures that further improve YOLOv4 object detection performance when tested on challenging target data. These architectures are: (a) Progressive Feature Reduction, (b) Unified Classifier, and (c) Integrated MS-DAYOLO framework that combines the benefits of the other two architectures. The corresponding DAN networks for these architectures are explained in detail later in this paper.
    \item We conducted extensive experiments using the Cityscapes, KITTI, and Waymo datasets. These experiments show that our proposed MS-DAYOLO framework provides significant improvements to the performance of YOLOv4 when tested on the target domain. We also show that MS-DAYOLO provides comparable and sometimes superior object detection performance relative to state-of-the-art approaches that are based on Faster R-CNN object detector. As mentioned above, our proposed MS-DAYOLO architectures achieve this level of state-of-the-art performance while providing an order of magnitude improvements in terms of computational complexity when compared to Faster R-CNN based solutions. 
\end{enumerate}

The remainder of this paper is organized as follows. Section II briefly highlights related work. Section III describes the proposed multiscale domain adaptive framework for YOLO object detector in detail, including the novel architectures for domain adaptive network. Section IV provides experimental results of the proposed framework with analysis and discussion.  Finally, section V concludes the paper with a summary of the key findings of our work.

\section{Related Work}
In general, state-of-the-art CNN based object detection models can be classified into two groups: one-stage and two-stage based methods. One-stage object detectors predict bounding boxes of objects and class probabilities associated with these objects directly from a full image in single computation via a unified neural network. The most well-known models of one-stage object detectors are YOLO \cite{YOLO,yolov2,yolov3,yolov4}, SSD \cite{liu2016ssd}, and RetinaNet \cite{RetinaNet}. On the other hand, two-stage object detectors generate proposal bounding boxes that potentially have an object using Region Proposal Network (RPN) in the first stage. Then, the proposals are fed to a second stage where cropped features are used to classify objects and fine-tune the bounding boxes. The most well-known models of two-stage object detectors are the R-CNN \cite{girshick2014rich} series, including Fast R-CNN \cite{girshick2015fast}, Faster R-CNN \cite{FasterRCNN}, R-FCN \cite{R-FCN}, and Mask R-CNN \cite{mask-RCNN}.         
 
 Recently, unsupervised domain adaptation has been used to improve the performance of object detection due to domain shift \cite{survey}. It attempts to learn a robust object detector using labeled data from the source domain and unlabeled data from the target domain. Domain adaptation approaches for object detection can mainly be classified into reconstruction based and adversarial-based solutions \cite{survey}. Reconstruction based domain adaptation attempts to improve the performance of an object detector for target domain by using image-to-image translation models \cite{arruda2019cross, lin2019cross,guo2019domain,devaguptapu2019borrow,inoue2018cross}. In particular, it utilizes image-to-image translation methods to generate artificial (fake) samples of the target domain from the corresponding source labeled samples. Consequently, translating labeled source data into corresponding target data will help in the training of an object detector in a target domain; and this should improve the performance of object detection in that domain. 
 
  In adversarial-based, a domain discriminator is trained to classify whether a data point is from the source or target domain, while the feature extractor of the object detector is trained to confuse the domain discriminator \cite{GANs}. Consequently, the feature extractor generates domain invariant features as a result of this training strategy. Many adversarial-based domain adaptation methods have been proposed for the Faster R-CNN object detector \cite{chen2018domain, zhu2019adapting,wang2019few,saito2019strong,he2019multi,sindagi2020prior, ifan_AAAI2020,CT2020,PDA2020}. The state-of-the-art approach of adversarial-based domain adaptation is Domain Adaptive Faster R-CNN \cite{chen2018domain}. Subsequently, many other approaches were proposed. For example, He and Zhang \cite{he2019multi} proposed multiple adversarial submodules for both domain and proposal features alignment. Furthermore, Zhao \textit{et al.} \cite{CT2020} proposed a collaborative self-training method that can propagate the loss gradient through the whole detection network, and mutually enhance the region proposal network and the region proposal classifier. In addition, Xu \textit{et al.} \cite{PDA2020} utilized elaborate prototype representations to achieve category-level domain alignment. On the other hand, Zhang \textit{et al.} \cite{zhang2021domain} applied domain adaptation modules in \cite{chen2018domain} to YOLOv4 object detector \cite{yolov3}.
 
 It is worth noting that most previous approaches for domain adaptation object detection, used Faster R-CNN as the base detector. Despite its popularity, Faster R-CNN suffers from a long inference time to detect objects. As a result, it is arguably not the optimal choice for time-critical, real-time applications such as autonomous driving. On the other hand, one-stage object detectors, and in particular YOLO, can operate quite fast, even much faster than real-time, and this makes them invaluable for autonomous driving and similar time-critical applications. Besides the computational advantage of the YOLO detector, YOLOv4 has many salient improvements and its object detection performance has improved rather significantly relative to prior YOLO architectures and more important in comparison to Faster R-CNN. All of these factors motivated our focus on the development of a new domain adaptation framework for YOLOv4.

\section{Proposed MultiScale Domain Adaptive YOLO}
\begin{figure*}[!t]
\begin{center}
\includegraphics[width=1\linewidth]{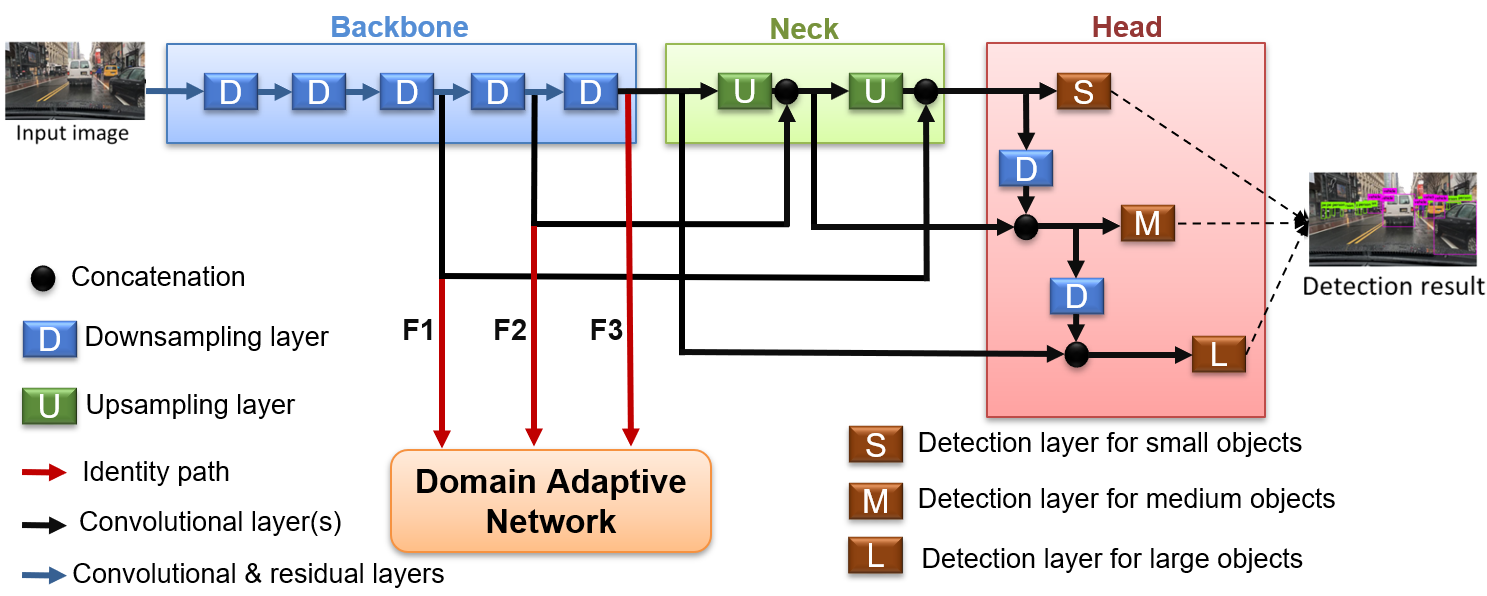} 
\end{center}
 \caption{Architecture of YOLOv4 with domain adaptation network (DAN) to develop domain adaptive YOLO. The details architectures of DAN are shown in Figure \ref{fig:DAN-archs}.}
\label{fig:arch_yolo}
\end{figure*}

YOLOv4\cite{yolov4}  incorporates many new revisions and novel techniques to improve the overall detection accuracy relative to its predecessor. YOLOv4 has three main parts: backbone, neck, and head as shown in Figure \ref{fig:arch_yolo}. The backbone is responsible for extracting multiple layers of features at different scales. The neck collects these features from three different scales of the backbone using upsampling layers and feeds them to the head. Finally, the head predicts bounding boxes surrounding objects as well as class probabilities associated with each bounding box.
 
The backbone (\textit{i.e.} {feature extractor}) represents a major module of the YOLOv4 architecture, and we believe that it makes a significant impact on the overall performance of the detector. In addition to many convolutional layers, it has 23 residual blocks \cite{ResNet}, and five downsampling layers to extract critical layers of features that are used by the subsequent detection stages. Here, we concentrate on the features that are fed to the neck module (F1, F2, and F3 in Figures \ref{fig:arch_yolo}). In particular, our goal is to apply domain adaptation to these three features to make them robust against domain shifts, and hence, have them converge toward domain invariance during domain-adaptation based training. Equally important, these three stages of features have different dimensions due to the successive downsampling layers that progressively reduce the width and height of features by half while doubling the number of channels. If $d$ is the width of the feature at the first scale (F1), then the dimensions of the three stages of features are: F1: $d \times d\times 256$, F2: $\frac{d}{2} \times \frac{d}{2} \times 512$, and F3: $\frac{d}{4} \times \frac{d}{4} \times 1024$.  

\subsection{Domain Adaptive Network for YOLO}
The proposed Domain Adaptive Network (DAN) is attached to the YOLOv4 object detector only during training in order to learn domain invariant features. Indeed, YOLOv4 and DAN are trained in an end-to-end fashion. For inference, and during testing, domain-adaptive trained weights are used in the original YOLOv4 architecture (without the DAN network). Therefore, our proposed framework will not increase the underlying detector complexity during inference, which is an essential factor for many real-time applications such as autonomous driving.

DAN uses the three distinct scale features of the backbone that are fed to the neck as inputs. It has several convolutional layers to predict the domain class (either source or target). Then, domain classification loss ($\mathcal{L}_{dc}$) is computed via binary cross entropy as follows:

\begin{equation}
\label{dc_loss}
 \mathcal{L}_{dc} = -\frac{1}{N}\sum_{i,x,y} [ t_i \ln p_i^{(x,y)} + (1-t_i) \ln (1-p_i^{(x,y)})]
 \end{equation}
 
Here, $t_i$ is the ground truth domain label for the i-th training image, with $t_i=1$ for source domain and $t_i=0$ for target domain. $p_i^{(x,y)}$ is the predicted domain class probabilities for i-th training image at location $(x,y)$ of the feature map. $N$ represents the total number of images in a batch multiplied by the total number of elements in the feature map.         

DAN is optimized to differentiate between the source and target domains by minimizing this loss. On the other hand, the backbone is optimized to maximize the loss to learn domain invariant features. 
Thus, features of the backbone should be indistinguishable for the two domains. Consequently, this should improve the performance of object detection for the target domain. 

To solve the joint minimization and maximization problem, we employ the adversarial learning strategy \cite{GANs}. In particular, we achieve this contradictory objective by using a Gradient Reversal Layer (GRL) \cite{GRL,ganin2016domain} between the backbone and the DAN network. GRL is a bidirectional operator that is used to realize two different optimization objectives. In the feed-forward direction, the GRL acts as an identity operator. This leads to the standard objective of minimizing the classification error when performing local backpropagation within DAN. On the other hand, for backpropagation toward the backbone network, the GRL becomes a negative scalar ($\lambda$). Hence, in this case, it leads to maximizing the binary-classification error; and this maximization promotes the generation of domain-invariant features by the backbone.

To compute the detection loss ($\mathcal{L}_{det}$) \cite{yolov4}, only source images are used because they are annotated with ground-truth objects. Consequently, all three parts of YOLOv4 (\textit{i.e.} backbone, neck and head) are optimized via minimizing $\mathcal{L}_{det}$. On the other hand, both source labeled images and target unlabeled images are used to compute the domain classification loss  ($\mathcal{L}_{dc}$) which is used to optimize DAN via minimizing it, and the backbone via maximizing it. As a result, both $\mathcal{L}_{det}$ and $\mathcal{L}_{dc}$ are used to optimize the backbone. In other words, the backbone is optimized by minimizing the following total lose:
\begin{equation}
\label{loss_backnone}
 \mathcal{L}_t = \mathcal{L}_{det} + \lambda \mathcal{L}_{dc}
 \end{equation}
 where $\lambda$ is a negative scalar of GRL that balances a trade-off between the detection loss and domain classification loss. In fact, $\lambda$ controls the impact of DAN on the backbone.    

\subsection{DAN Architectures}
\label{sec:DAN_Arch}
\begin{figure}
\begin{center}
  \includegraphics[width=1\linewidth]{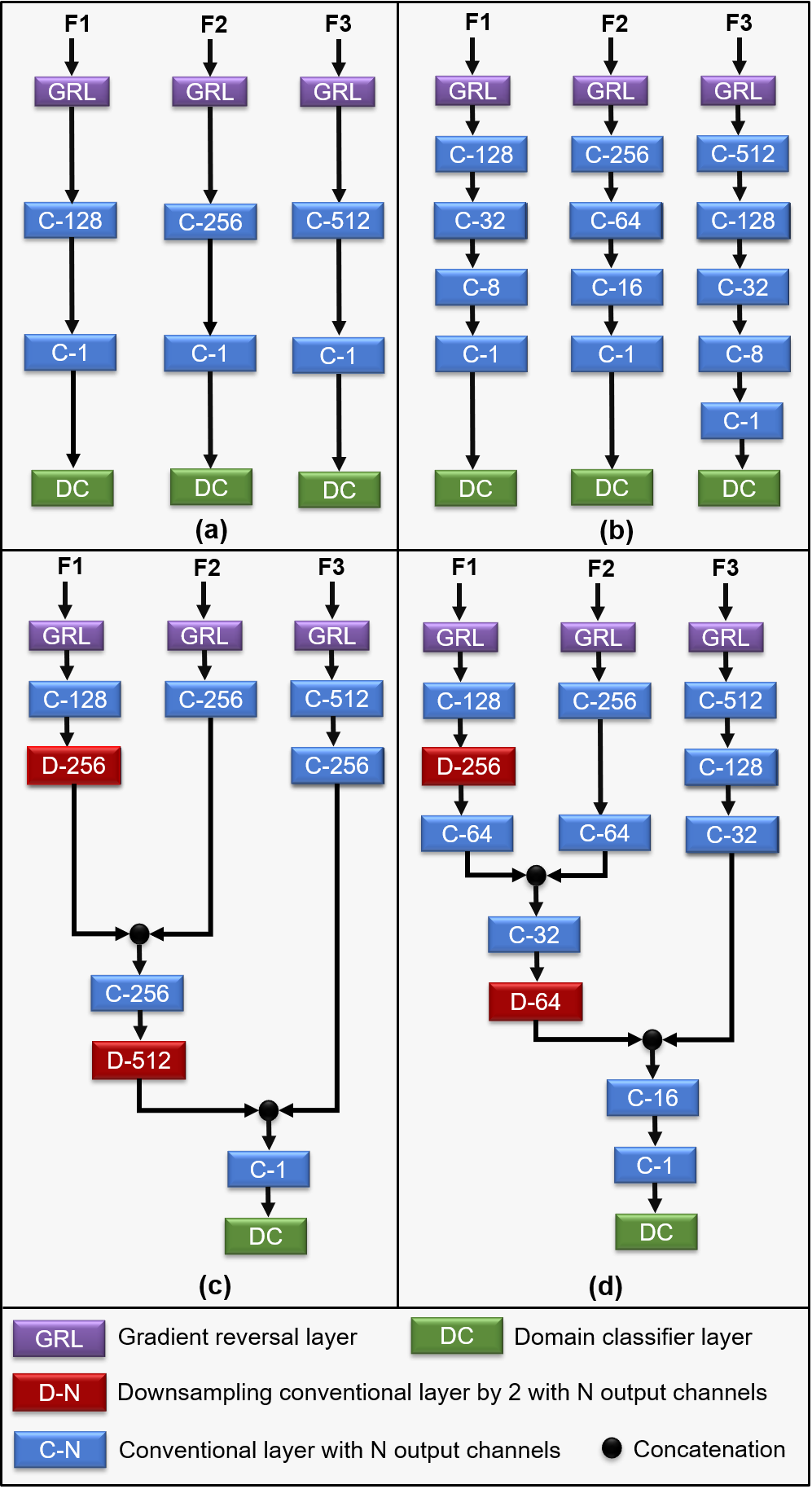}
\end{center}
  \caption{Proposed architectures for the Domain Adaptive Network (DAN): (a) Baseline, (b) Progressive Feature Reduction (PFR), (c) Unified Classifier (UC), (d) Integrated. F1, F2, and F3 are features of the backbone network that are fed to the neck.}
\label{fig:DAN-archs}
\end{figure}

We developed various architectures for the Domain Adaptive Network (DAN) as shown in Figure \ref{fig:DAN-archs} to explore and gain insight into the impact of different components on achieving improved performance for the target domain. Under all of our architectures, we employ a multiscale strategy that connects the three features F1, F2, and F3 of the backbone to the DAN through three corresponding GRLs. Other than this common multiscale strategy, the proposed DAN architectures differ from each other as explained below.

\noindent \textbf{a- Multiscale Baseline }: Instead of applying domain adaptation for only the final scale of the feature extractor as done in the Domain Adaptive Faster R-CNN architecture \cite{chen2018domain}, we develop domain adaptation for three scales separately to solve the gradient vanishing problem. In other words, applying domain adaptation only to the final scale (F3) does not make a significant impact on the previous scales (F1 and F2) due to the gradient vanishing problem as there are many layers between them. As a result, we apply domain adaptation to all scales as shown in Figure \ref{fig:DAN-archs} (a). For each scale, there are two convolutional layers after GRL, the first one reduces the feature channels by half, and the second one predicts the domain class probabilities. Finally, a domain classifier layer is used to compute the domain classification loss. 

\noindent  \textbf{b- Progressive Feature Reduction (PFR)}: As shown in Figure \ref{fig:DAN-archs}(a), the baseline architecture reduces the feature vector size resulting from the YOLOv4 backbone into a single-value feature (scalar) rather abruptly through two stages of neural networks. This simple two-stage DAN aims at generating a single feature value that serves as an input into the domain classifier. The fact that the domain classifier requires a single feature value is inherent in the binary nature of the classifier that simply needs to classify the image data into either source domain or target domain.  Meanwhile, and due to the adversarial strategy used here, the above baseline domain adaptation network is competing with the significantly more complex network of the backbone as shown in Figure \ref{fig:arch_yolo}. We observed that this mismatch between the simplistic baseline DAN architecture and the complex backbone network could compromise the domain adaptation performance. Thus, the DAN network may not be sufficiently powerful to distinguish between the source and target domains since the complex backbone network can easily confuse (and trick) the DAN network. To mitigate this mismatch, we increase the number of convolutional layers for each scale by progressively reducing the feature channels as shown in Figure \ref{fig:DAN-archs}(b). This progressive reduction of feature channels helps the DAN network to compete more efficiently against the more complex backbone. As a result, the extracted features by the backbone network will be more domain invariant.

Therefore, while the baseline architecture reduces the number of the feature channels using two stages of neural networks, our proposed progressive-feature-reduction employs four or five stages depending on the original feature size. In particular, for feature vectors F1 and F2, we employ four stages of neural networks that progressively reduce the feature vector size from 128 and 256, respectively, toward a single-feature scalar value, which is all that is required as an input for the binary domain classifier. For feature vector F3, we employ a five-stage neural network DAN to progressively reduce the backbone feature vector toward the scalar feature value. It is important to highlight the following regarding the proposed progressive-feature-reduction architecture. It is possible to employ a larger number of stages of progressive reduction than the number of stages we employed in our architecture shown in Figure \ref{fig:DAN-archs}(b). However, based on our experience, increasing the number of stages beyond four or five stages does not necessarily improve the overall performance. 

\noindent  \textbf{c- Unified Classifier (UC)}: Under the multiscale baseline and Progressive Feature Reduction architectures, each scale has its own distinct domain classifier. This multi-classifier strategy may lead to inconsistency among scales. For example, a domain classifier at one scale may classify an image patch as a source data, while a domain classifier at another scale may classify the same image patch as originating from the target domain. (Examples of this inconsistency are shown later in the Experiments section.) To address this potential inconsistency, we propose to use a single (unified) domain classifier that combines the feature vectors from all scales as shown in Figure \ref{fig:DAN-archs}(c). It is important to highlight the following about the proposed Unified Classifier (UC) domain adaptive network:
\begin{enumerate}
\item We use downsampling convolutional layers to match the size of features at different scales. For example, in order to combine the feature vectors resulting from the F1 and F2 scales of the backbone, we add a downsampling stage to the F1 scale and concatenate the resulting vector with the feature vector from F2. This strategy maintains the multiscale attribute of our domain adaptive network while targeting a unified domain classifier architecture.
\item Furthermore, we concatenate features at different scales in a way to make each scale contributes equally in terms of the number of feature channels. In other words, each feature scale equally contributes in the prediction of the domain class probabilities. 
\end{enumerate}          

\noindent  \textbf{d- Integrated}: It is important to note that the above two improvements,  progressive feature reduction (PFR) and unified classifier (UC), have been applied directly and separately to the multiscale baseline architecture. Consequently, to gain the benefits of both, the progressive feature reduction and unified domain classifier strategies, we integrate them in one network as shown in Figure \ref{fig:DAN-archs}(d). In principle, we have developed the network by complementing the unified-classifier architecture (Figure \ref{fig:DAN-archs}(c)) with additional stages of convolutional layers to achieve a more progressive reduction in feature channel sizes. This is evident by comparing the two architectures shown in Figures \ref{fig:DAN-archs}(c) and \ref{fig:DAN-archs}(d).

\section{Experiments}
In this section, we evaluate our proposed domain adaptive YOLO framework and the proposed MS-DAYOLO architectures. We modified the official source code of YOLOv4 that is based on the darknet platform\footnote{https://github.com/AlexeyAB/darknet}, and we developed a new code to implement our proposed methods\footnote{https://github.com/Mazin-Hnewa/MS-DAYOLO}.

\subsection{Setup}
For training, we used the default settings and hyper-parameters that were used in the original YOLOv4\cite{yolov4}. The network is initialized using the pre-trained weights file. The training data includes two sets: source data that has images and their annotations (bounding boxes and object classes), and target data without annotation. Each batch has 64 images, 32 from the source domain and 32 from the target domain. Based on prior works \cite{chen2018domain, zhu2019adapting,he2019multi} and on our experience using trial and error, we set $\lambda =0.1$ for all experiments. 

For evaluation, we report Average Precision (AP) for each class as well as mean average precision (mAP) with a threshold of 0.5 \cite{2011pascal} using testing data that has labeled images of the target domain. We have followed other prior domain adaptive object-detection works that use the same threshold value of 0.5. We compare our proposed method with the original YOLOv4 and other state-of-the-art domain adaptation approaches that are based on Faster R-CNN object detector \cite{FasterRCNN}, all applied to the same target domain validation set.

\subsection{Results}
\label{res}

\subsubsection{Cross Camera Adaptation}
Domain shift can occur between different real visual datasets captured by different driving vehicles equipped with different cameras even if these visuals are taken under similar weather conditions. Such domain shift is usually driven by different camera setups leading to a shift in image quality and resolution. Moreover, such datasets are usually captured in various locations, which have different views and driving environments. All these factors lead to domain disparity between datasets. Under this experiment, we evaluate the performance of our MS-DAYOLO framework for domain adaptation between two real driving datasets: KITTI \cite{kitti} and Cityscapes as has been done by many recent works in this area \cite{chen2018domain, he2019multi,khodabandeh2019robust,zhu2019adapting,PDA2020}. In particular, the KITTI training set which has 6000 labeled images, is utilized as source data. While the Cityscapes training set which has 2975 images, but without labels, is utilized as target data. The Cityscapes validation set which has 500 labeled images, is used for testing and evaluation. 

\begin{figure}
\begin{center}
  \includegraphics[width=1\linewidth]{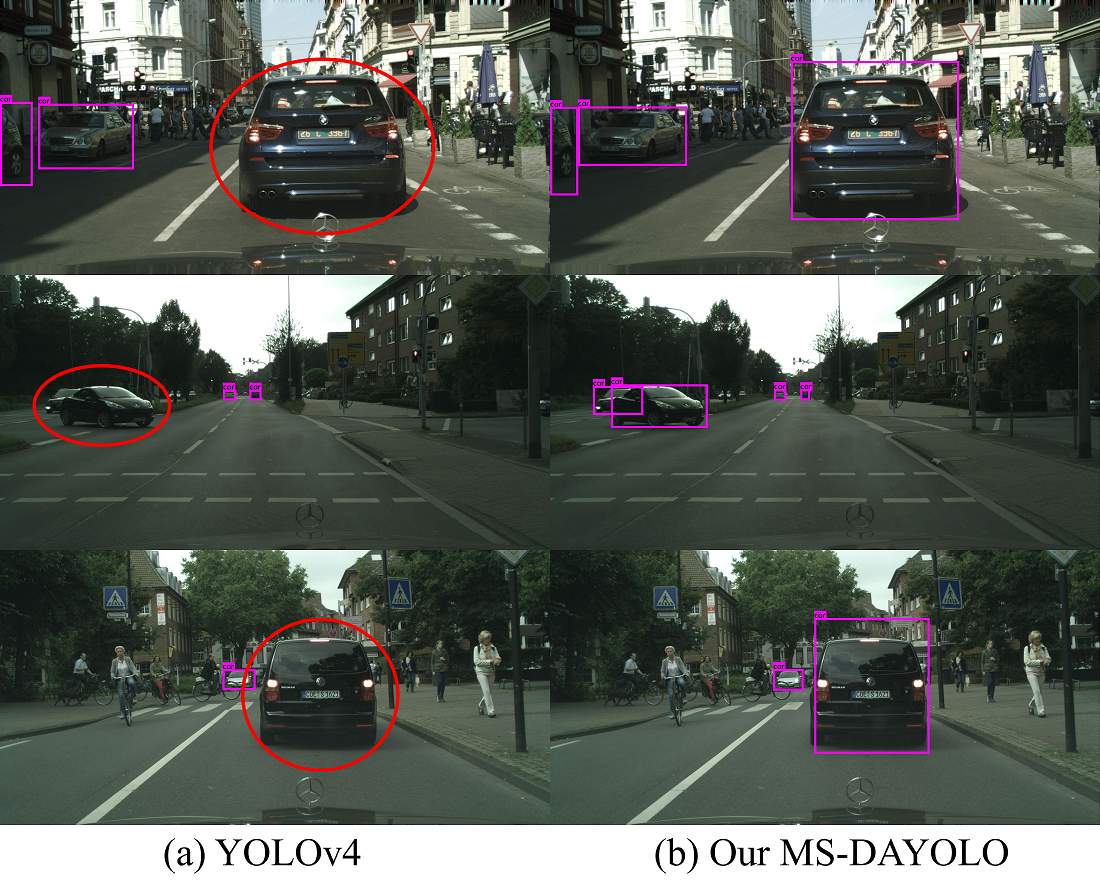}
\end{center}
  \caption{Visual detection examples of the KITTI $\rightarrow$ Cityscapes experiment for the car class using (a) the original YOLOv4, and (b) Our proposed integrated MS-DAYOLO applied onto the Cityscapes validation set. These examples show that the integrated MS-DAYOLO successfully detects the vehicles in the scenes while the original YOLOv4 fails to detect the same vehicles.}
\label{fig:k2c_examples}
\end{figure}

Table \ref{mAP_KITTI_Cityscapes} presents the performance results based on the car AP as has been reported by prior works \cite{khodabandeh2019robust,chen2018domain, zhu2019adapting,he2019multi,PDA2020} because it is the only common object class between the two datasets. Based on these results, all of the architectures of our proposed framework outperform the original YOLOv4 approach by a significant margin. Moreover, the proposed integrated architecture achieves the best overall performance in terms of mAP. Figure \ref{fig:k2c_examples} shows visual examples for qualitative comparison of our method with the original YOLOv4. It is obvious from these examples that our approach successfully detects the vehicles in the scenes while the original YOLOv4 fails to detect the same vehicles.

\begin{table}[!t]
\caption{Quantitative results of cross camera adaptation from KITTI to Cityscapes based on the car AP. The \textit{car} class was selected because it is the only common object class between the two datasets. The MS-DAYOLO uses YOLOv4 object detector \cite{yolov4}, while the other methods use Faster R-CNN object detector \cite{FasterRCNN}. The inference time is measured in Frames Per second (FPS) using NVIDIA GeForceGTX 1080 Ti GPU.}
\begin{center}
\label{mAP_KITTI_Cityscapes}
\begin{tabular} {|l|c|c|c|} 
\hline
Method & Backbone   & Car AP & FPS\\
\hline
DAF (CVPR'18)\cite{chen2018domain} & \multirow{5}{*}{VGG16} & 38.5 &  \multirow{5}{*}{6.2}\\
MAF (ICCV'19)\cite{he2019multi} &  & 41.0 & \\
CT (ECCV'20)\cite{CT2020} & & 43.6 & \\
PDA (WACV'20)\cite{PDA2020} &    & 43.9  & \\
MeGA (CVPR'21)\cite{mega} &  & 43.0 &\\
\hline
DSS (CVPR'21)\cite{dss} & ResNet50 & 42.7 & 3.6\\
\hline

\multicolumn{2}{|l|}{YOLOv4} &  44.5  & \multirow{5}{*} {\textbf{48.2}}  \\
\cline{1-3}

\multirow{4}{*} {MS-DAYOLO} &{Baseline} &  45.5  & \\
 & PFR   & 46.8 &  \\

& UC & 	47.3 &   \\
& Integrated & \textbf{47.6} &   \\
\hline
\end{tabular}
\end{center}
\end{table}

\subsubsection{Adverse Weather Adaptation}
Domain shift due to changes in weather conditions is one of the most prominent reasons for the discrepancy between the source and target domains. Reliable object detection systems in different weather conditions are essential for many critical applications such as autonomous driving.
As a result, we focus on presenting the evaluation results of our proposed MS-DAYOLO framework by studying domain shifts under challenging weather conditions for autonomous driving. To achieve this, we use three different driving datasets: Cityscapes \cite{Cityscapes}, Foggy Cityscapes \cite{Foggy_cityscapes}, and Waymo \cite{waymo}.

\textbf{Clear $\rightarrow$ Foggy}: We discuss the ability of our proposed method to adapt from clear to foggy weather using driving datasets: Cityscapes \cite{Cityscapes} and Foggy Cityscapes \cite{Foggy_cityscapes} as has been done by many recent works in this area \cite{chen2018domain,sindagi2020prior,wang2019few,he2019multi,khodabandeh2019robust,cai2019exploring,zhu2019adapting,saito2019strong,kim2019diversify}. The Cityscapes training set has 2975 labeled images that are used as source domain. Similarly, the Foggy Cityscapes training set also has 2975 images, but without annotations, and are used as the target domain. Original YOLOv4 is trained using the source domain data only. While MS-DAYOLO is trained using both source and target domain data. The Foggy Cityscapes validation set has 500 labeled images which are used for testing and evaluation. Because the Foggy Cityscapes training set is annotated, we are able to train the original YOLOv4 with this set to show the ideal performance (oracle).

\begin{table*}[!t]

\caption{Quantitative results of domain adaptation for the clear $\rightarrow$ foggy experiment of the Cityscapes dataset. The MS-DAYOLO uses YOLOv4 object detector \cite{yolov4}, while the other methods use Faster R-CNN object detector \cite{FasterRCNN}. *The results are reported from \cite{cai2019exploring}. The inference time is measured in Frames Per Second (FPS) using NVIDIA GeForceGTX 1080 Ti GPU.}

\label{mAP_cityscapes}
\centering
\begin{tabular} {|l|c|cccccccc|c|c|}
\hline
Method & Backbone & Person & Rider
 & Car & Truck & Bus & Train & Mcycle & Bicycle &
 mAP&FPS  \\
\hline
DAF (CVPR'18)\cite{chen2018domain} & \multirow{7}{*}{VGG16} & 25.0 & 31.0 & 40.5 & 22.1 & 35.3 & 20.2 & 20.0 & 27.1 & 27.6 &  \multirow{7}{*}{6.2} \\

 MAF (ICCV'19)\cite{he2019multi} &  & 28.2 & 39.5 & 43.9 & 23.8 & 39.9 & 33.3 & 29.2 & 33.9 & 34.0 &  \\
 
iFAN (AAAI'20)\cite{ifan_AAAI2020} &  & 32.6 & 40.0 &48.5 & 27.9 & 45.5 & 31.7  & 22.8 & 33.0    & 35.3 &  \\
CT (ECCV'20)\cite{CT2020} & & 32.7 & 44.4 & 50.1 & 21.7 & 45.6 & 25.4 &  30.1 & 36.8 & 35.9 &  \\

PDA (WACV'20)\cite{PDA2020} &  & 36.0 & 45.5 & 54.4 & 24.3 & 44.1 & 25.8 & 29.1 & 35.9 & 36.9 &  \\
ECR (CVPR'20)\cite{xu2020exploring} &  & 32.9 & 43.8 & 49.2 & 27.2 & 45.1 & 36.4 & 30.3 & 34.6 & 37.4 &  \\
MeGA (CVPR'21)\cite{mega} &  & 37.7 & 49.0 & 52.4 & 25.4 & 49.2 & \textbf{46.9} & 34.5 & 39.0 & \textbf{41.8} &  \\
\hline
DAF (CVPR'18)\cite{chen2018domain}*  &  \multirow{4}{*} {ResNet50} & 29.2 & 40.4 & 43.4 & 19.7 & 38.3 & 28.5  & 23.7 & 32.7 & 32.0 &  \multirow{4}{*}{3.7} \\

MTOR (CVPR'19)\cite{cai2019exploring} &    & 30.6 &41.4 & 44.0 & 21.9 & 38.6 & 40.6 & 28.3 & 35.6 & 35.1  &  \\
DSS (CVPR'21)\cite{dss} &    & \textbf{42.9} & \textbf{51.2} & 53.6 & \textbf{33.6} & 49.2 & 18.9 & \textbf{36.2} & \textbf{41.8} & 40.9 &  \\
\hline
\multicolumn{2}{|l|}{YOLOv4} & 31.6 & 38.3 & 46.9	&23.9	& 39.9	&20.1	& 16.8	&30.3 & 31.0 &  \multirow{6}{*}{\textbf{48.2}} \\
\cline{1-11}
\multirow{4}{*}{MS-DAYOLO}  & Baseline & 38.6	 &45.5&	55.9&	22.8&	45.6&	32.5&	28.8&	36.5&	38.3 & \\
 & PFR   & 38.5 &	46.5 &	56.5 &	27.6 &	48.7 &	38.5 &	26.4 &	38.4 &	 40.1  &\\
& UC & 39.3 &	45.0 &	\textbf{57.0} & 29.9 &	48.0 &	36.6 &	30.2 &	36.4 &	 40.3 &   \\

& Integrated & 39.6 &	46.5 &	56.5 &	28.9 &	\textbf{51.0} &	45.9  &	27.5 &	36.0 &	 41.5 &   \\
\cline{1-11}
 \multicolumn{2}{|l|}{YOLOv4 trained with target (Oracle)} & 42.4 & 49.5  &	63.6  &	37.6  &	59.8 &	47.1 &	31.1 &	39.9 &	46.3 &   \\
\hline
\end{tabular}
\end{table*}

\begin{figure}[t]
\begin{center}
   \includegraphics[width=1\linewidth]{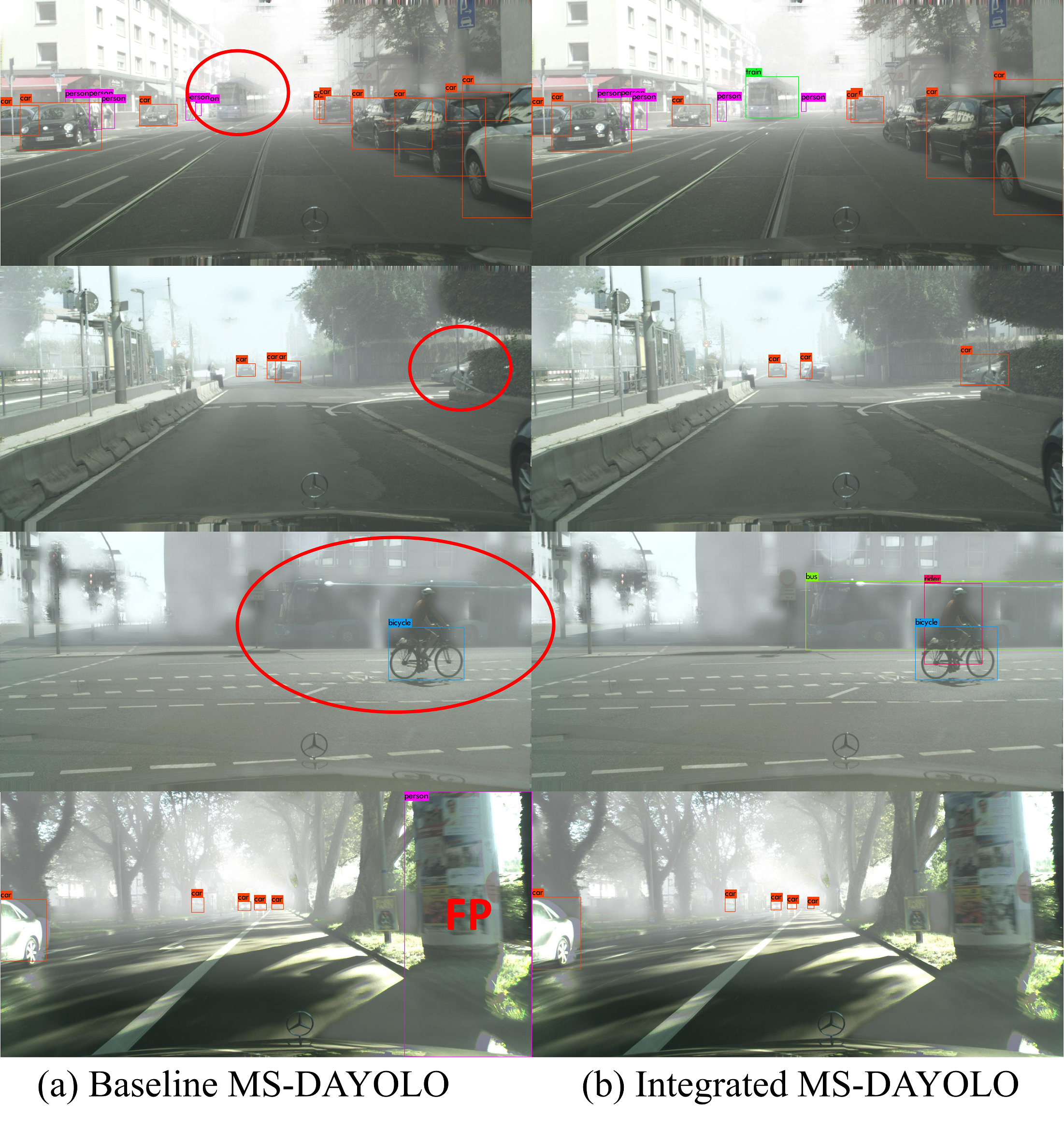}
\end{center}
  \caption{Visual detection examples of the clear $\rightarrow$ foggy experiment using (a) the baseline architecture, and (b) the integrated architecture of MS-DAYOLO applied onto the foggy images extracted from the Foggy Cityscapes datasete \cite{Foggy_cityscapes}. The three top examples show that the integrated architecture successfully detects particular objects that the baseline architecture fails to detect. In the bottom example, the baseline suffers from a false positive problem, while the integrated mitigates this false positive. \textcolor{red}{\textbf{FP}}: implies false positive.}
\label{fig:cityscapes2}
\end{figure}

Table \ref{mAP_cityscapes} summarizes the performance results. A clear performance improvement is achieved by our method over the original YOLOv4. We also observe that the proposed Progressive Feature Reduction (PFR), Unified Classifier (UC), and Integrated architectures improve the detection performance relative to the baseline architecture. Although the MeGA method outperforms Integrated MS-DAYOLO by a small margin (0.3\%), our MS-DAYOLO runs faster than real-time, and it is significantly faster than GPA in terms of frames per second (FPS), which is essential for time-critical applications. It is worth noting that the proposed integrated architecture achieves significant improvements relative to the original YOLOv4, and it almost reaches the performance of the ideal (oracle) scenario, especially for some object classes in terms of average precision. Figure \ref{fig:cityscapes_examples} shows examples of detection results of the proposed method as compared to the original YOLOv4. Moreover, Figure \ref{fig:cityscapes2} shows examples of detection results of the proposed integrated architecture as compared to the baseline one.

\begin{figure}[t]
  \includegraphics[width=1\linewidth]{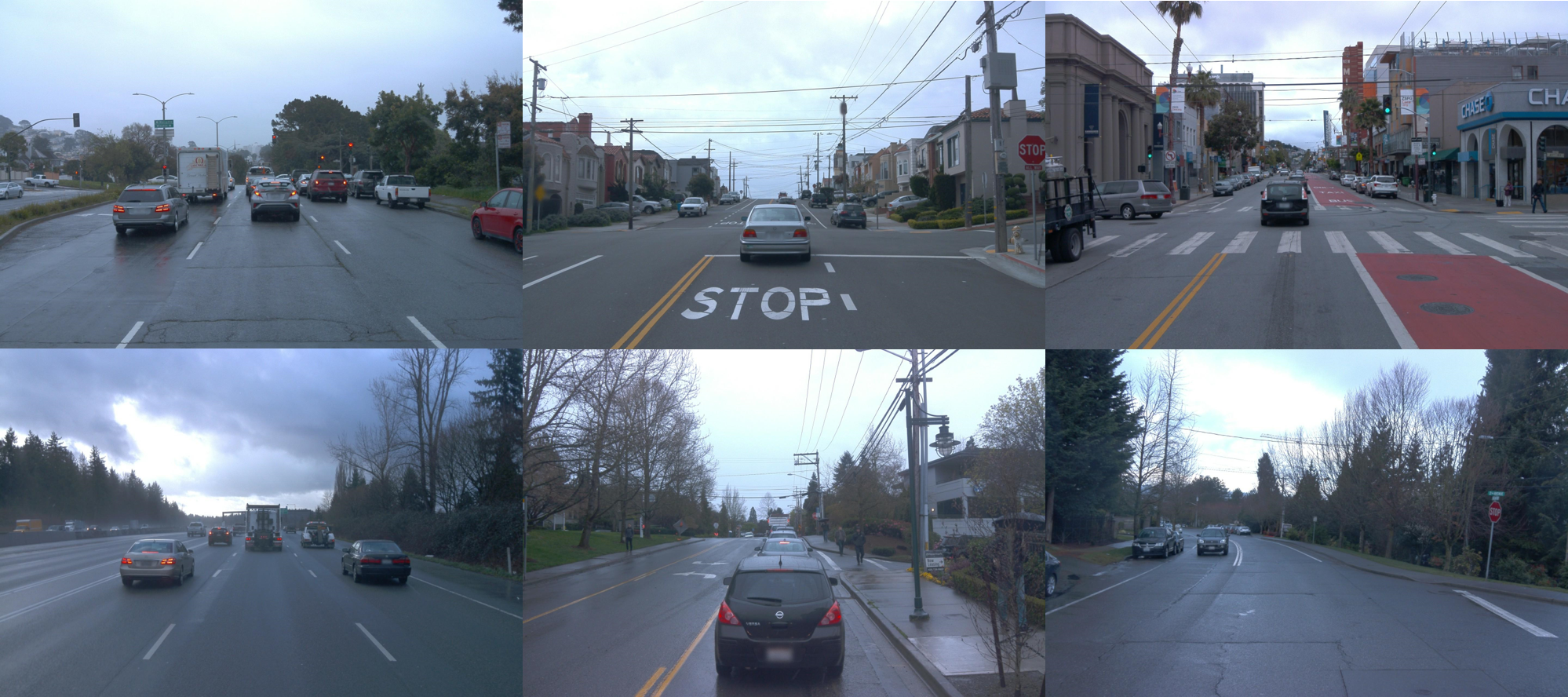}
  \caption{Examples of training images of Waymo dataset \cite{waymo}. Images in the top row are tagged as being captured in sunny weather, while images in the bottom row are tagged as being captured in rainy weather. It is obvious that the domain shift between sunny and rainy images is not very significant.} 
\label{fig:sunny_vs_rainy}
\end{figure}

\textbf{Sunny $\rightarrow$ Rainy}: we present results for applying YOLOv4 and our MS-DAYOLO framework on the Waymo dataset \cite{waymo}, which includes two sets of visual data that are designated as "sunny" and "rainy". We extracted 14319 "sunny weather" labeled images for the source data, and 13004 "rainy weather" unlabeled images to represent the target data. As before, the original YOLOv4 is trained using only source data (\textit{i.e.} labeled sunny images). Meanwhile, our proposed MS-DAYOLO is trained using both source and target data (\textit{i.e.} labeled sunny images and unlabeled rainy images). In addition, we extracted 1676 labeled images from the rainy-weather data for testing and evaluation. It is important to note the following key observations regarding the Waymo datasets: (a) The designations "sunny" and "rainy" images have been determined by the providers of the Waymo dataset. (b) From our extensive experience in working with this data, the distinction between sunny and rainy image samples is quite subjective, and in many cases, one can argue that a "rainy" image sample should be labeled as "sunny" or vice versa. Consequently, the domain shift between the two domains, which are designated as "sunny" and "rainy", is not very significant as shown in the examples of Figure \ref{fig:sunny_vs_rainy}. This is crucial since training the original YOLOv4 using the Waymo "sunny" dataset effectively covers a large number of "rainy" testing samples that fall within the source "sunny" domain. Nevertheless, we opted to follow the dataset designations with the aim of evaluating any potential improvements that the proposed MS-DAYOLO framework may provide.

\begin{table}[!t]
\centering
\caption{Quantitative results of domain adaptation for the sunny $\rightarrow$ rainy experiment of the Waymo dataset.}

\label{mAP_waymo}
\begin{tabular} {|l|c|cc|c|} 
\hline
\multicolumn{2}{|l|}{Method} & Person & Vehicle &
 mAP  \\
\hline
\multicolumn{2}{|l|}{YOLOv4} & 38.6 & 55.4 &  47.0 	 \\
\hline
\multirow{4}{*} {MS-DAYOLO } & Baseline   & 40.0 &	55.4 &	 47.7 \\
\cline{2-5}
& PFR   & 39.8  & 56.3 & 48.1  \\
\cline{2-5}
& UC & 39.4 &	56.7   &	48.0   \\
\cline{2-5}

& Integrated & \textbf{40.0} &	 \textbf{57.0}  & \textbf{48.5}	 \\
\hline
\end{tabular}
\end{table}

\begin{figure}
\begin{center}
  \includegraphics[width=1\linewidth]{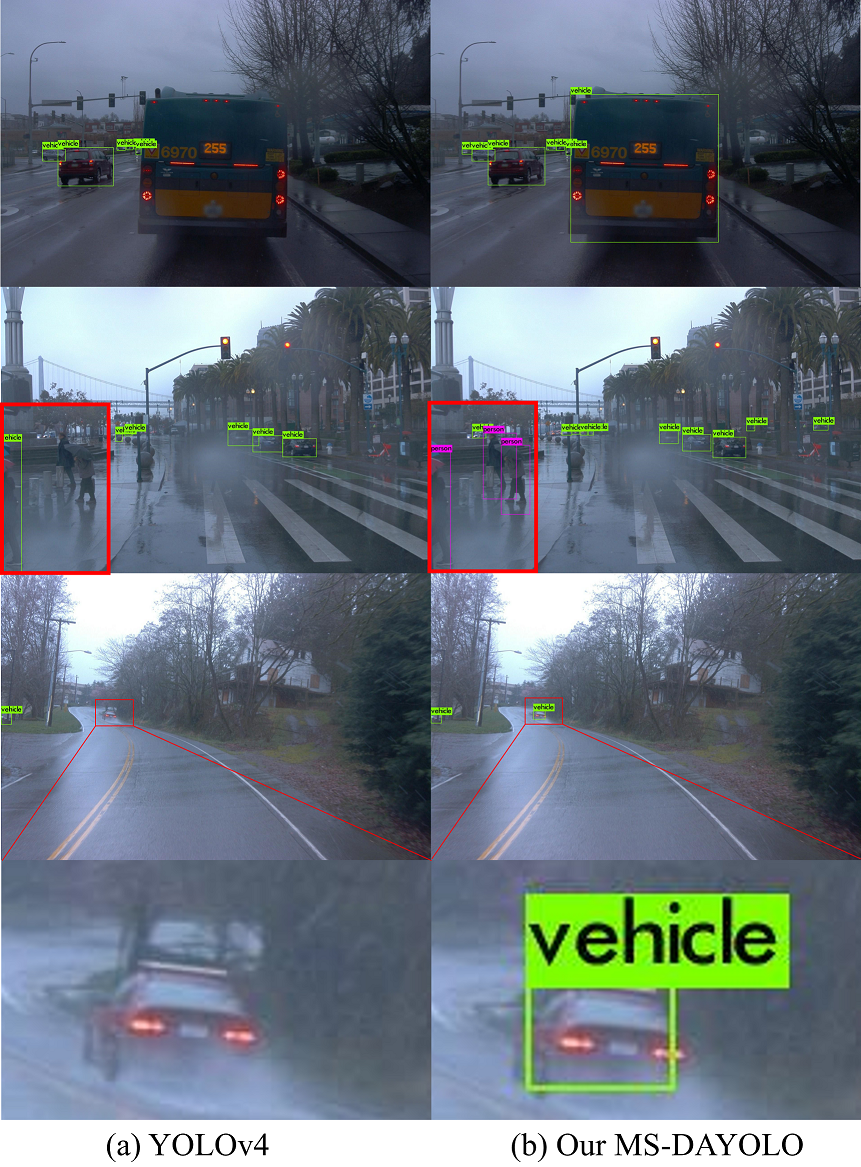}
\end{center}
  \caption{Visual detection examples of the sunny $\rightarrow$ rainy experiment using (a) the original YOLOv4, and (b) Our proposed Integrated MS-DAYOLO applied onto labeled rainy images extracted from the Waymo dataset \cite{waymo}. The green bounding box refers to the vehicle class while the purple one refers to the person class.}
\label{fig:waymo_examples}
\end{figure}

\begin{figure}
\begin{center}
  \includegraphics[width=1\linewidth]{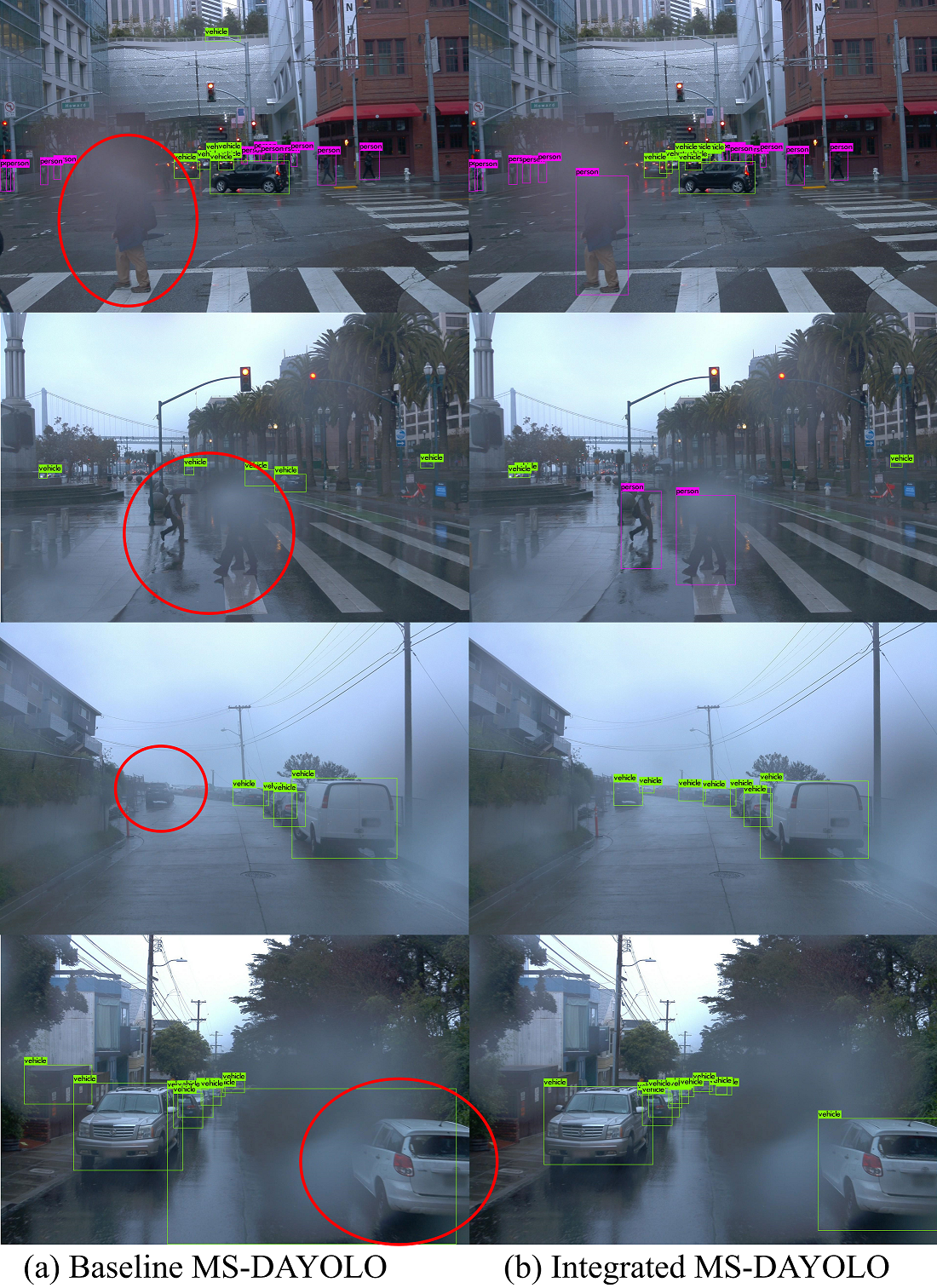}
 \end{center}
  \caption{Visual detection examples of the sunny $\rightarrow$ rainy experiment using (a) the baseline architecture, and (b) the integrated architecture of MS-DAYOLO applied onto the rainy images extracted from the Waymo dataset \cite{waymo}. The baseline MS-DAYOLO fails to detect pedestrians crossing the street in the top two images, and cars in the bottom two images, while the integrated MS-DAYOLO successfully detects these objects. The green bounding box refers to the vehicle class while the purple one refers to the person class.}
\label{fig:waymo_examples2}
\end{figure}

\begin{figure}
\begin{center}
  \includegraphics[width=1\linewidth]{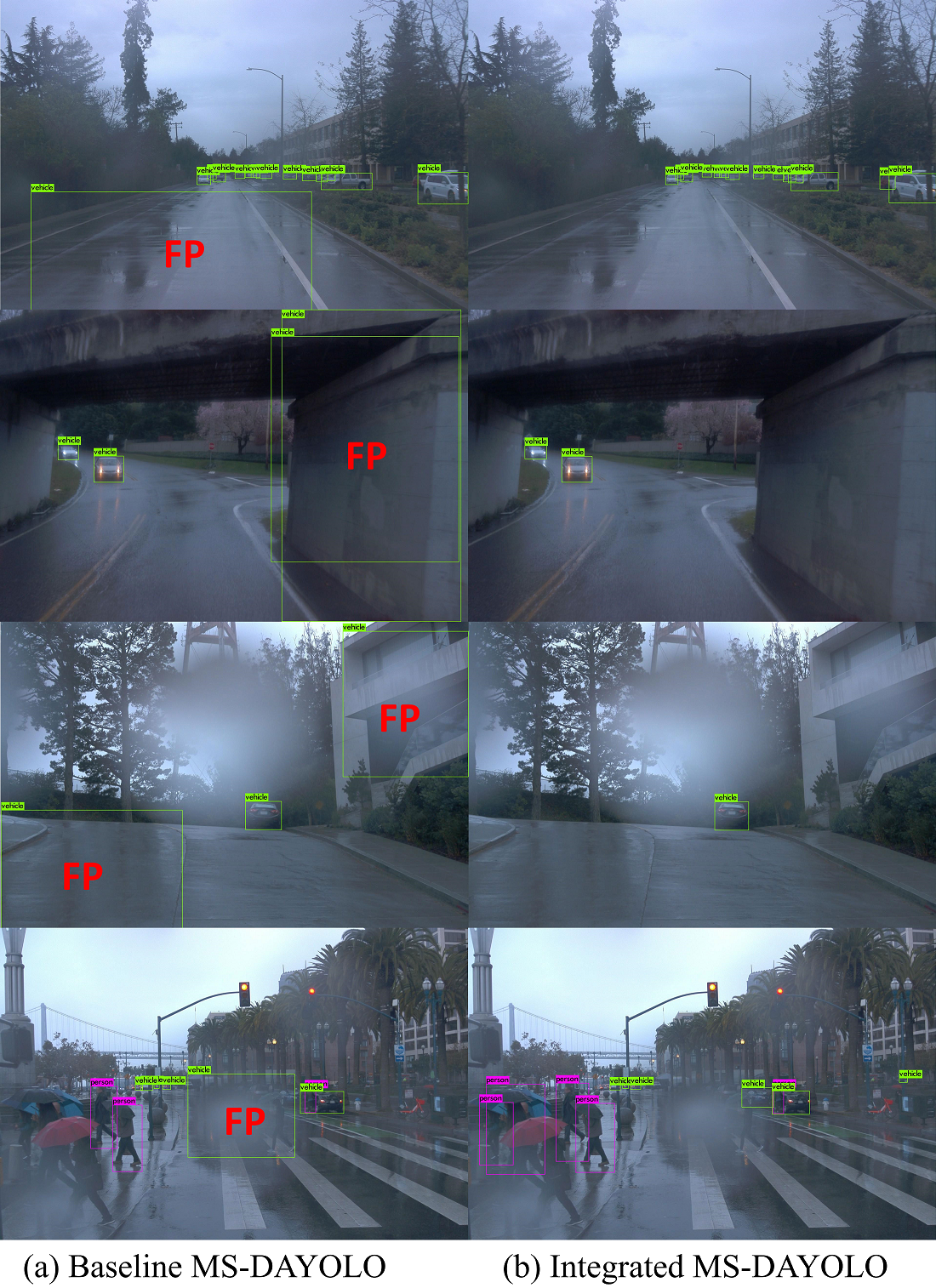}
 \end{center}
  \caption{Visual detection examples of the sunny $\rightarrow$ rainy experiment using (a) the baseline architecture, and (b) the integrated architecture of MS-DAYOLO applied onto the rainy images extracted from the Waymo dataset \cite{waymo}.  In these examples, the baseline MS-DAYOLO suffers from instances of false positive, while the integrated MS-DAYOLO eliminates these false positives. \textcolor{red}{\textbf{FP}}: means false positive. The green bounding box refers to the vehicle class while the purple one refers to the person class.}
\label{fig:waymo_examples_fp}
\end{figure}

The results are summarized in Table \ref{mAP_waymo}. It is clear that the MS-DAYOLO framework still provided good improvements despite the fact that, in this case, the two domains, sunny and rainy, have a significant overlap. This could explain why the improvements are not as salient as the improvements achieved when applying MS-DAYOLO on the Cityscapes data, which consisted of two clearly distinct domains as shown in the examples of Figure \ref{fig:cityscapes_examples}. Moreover, and similar to the clear $\rightarrow$ foggy experiment, we observe that the proposed Progressive Feature Reduction (PFR), Unified Classifier (UC), and Integrated architectures improve the detection performance relative to the baseline architectures when applied to the Waymo dataset. For this experiment, we do not report the performance of other domain adaptive object detection methods that are based on Faster R-CNN because none of these methods reported or used the Waymo dataset for the sunny $\rightarrow$ rainy domain-shift scenario. 
Figure \ref{fig:waymo_examples} shows examples of detection results of the proposed Integrated MS-DAYOLO framework as compared to the original YOLOv4. In addition, Figure \ref{fig:waymo_examples2} shows examples of detection results that the integrated architecture succeeds in detecting objects while the baseline architecture fails to detect the same objects. Furthermore, Figure \ref{fig:waymo_examples_fp} shows examples where the baseline architecture suffers from false positive cases, while the integrated one eliminates these false positives, which contribute to its improved performance.

\subsection{Ablation Study}

To show the importance of applying domain adaptation to three distinct scales of the backbone network, we conducted an ablation study for the clear $\rightarrow$ foggy experiment. First, we applied domain adaptation, separately, to each of the three scales of features that are fed into the neck of the YOLOv4 architecture. Also, we applied domain adaptation to different combinations of two scales at a time. Finally, we compared the results with the performance of applying these combinations of the study with the performance of applying our baseline MS-DAYOLO to all three scales as explained in section \ref{sec:DAN_Arch}.
Another important aspect of this ablation study is that we wanted to consider objects that have statistically significant numbers of sample data. In that context, because the number of ground-truth objects for some classes (truck, bus, and train) is small (\textit{i.e.} less than 500 in the training set, and 100 in the testing set), the performance measure will be inaccurate for these classes. As a result, we exclude them in this ablation study and compute mAP based on the remaining classes.

Table \ref{ablation} summarizes results of the ablation study. It is clear that based on these results, we can conclude that applying domain adaptation to all three feature scales improves the detection performance on the target domain, and achieves the best result.

\begin{table}[!t]
\centering
\caption{Ablation Study, \checkmark means that domain adaptation is applied to the feature scale(s) using our baseline MS-DAYOLO.}
\label{ablation}
\begin{tabular} {|c c c|c c c c c|c|} 
\hline
 F1 & F2 & F3 & Person & Rider
 & Car & Mcycle & Bicycle & mAP  \\
\hline
  &  &  & 31.6 & 38.3 & 46.9 &16.8	&30.3 & 32.8\\
\hline
 & & \checkmark & 36.8&	42.8&	53.7&		24.8&	32.4 & 38.1\\ 
\hline

 &\checkmark  &  &	37.1&	41.5 &	54.5&		26.2&	32.4 & 38.3\\
 
\hline
 \checkmark &  &  &	36.3 &	44.2 & 53.1 &		25.8&	35.9 & 39.1 \\

\hline

 & \checkmark& \checkmark & 36.6&	42.7&	55.7 &  26.1 &	33.5& 38.9  \\
\hline

\checkmark & \checkmark&  &37.5 &	42.5 &	54.5  &27.8 &	34.8 &	39.4 \\
\hline
\checkmark & & \checkmark &  36.4&	\textbf{46.1}&	52.2& 22.5 &35.0 &38.4\\
\hline

\checkmark & \checkmark& \checkmark &\textbf{38.6}	 &45.5&	\textbf{55.9} & \textbf{28.8} &	\textbf{36.5} &	\textbf{41.1}\\
\hline
\end{tabular}
\end{table}

\subsection{Analysis}
\begin{figure}
\begin{center}
  \includegraphics[width=1\linewidth]{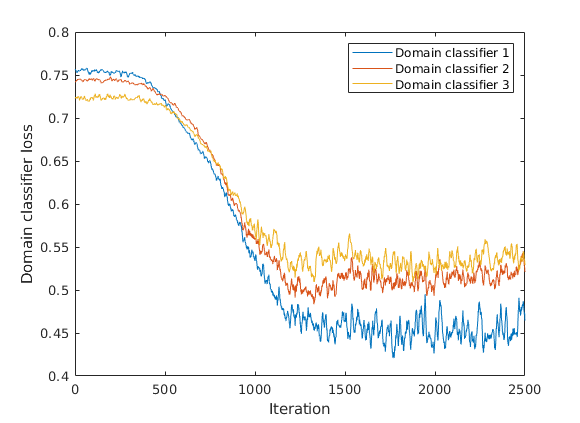}
\end{center}
  \caption{Losses of three domain classifiers of the baseline architecture over the first 2500 iterations of training.} 
\label{fig:3DCL}
\end{figure}

In order to show the benefit of using a unified domain classifier instead of three different domain classifiers, we recorded the domain classifier loss of Equation \ref{dc_loss} over training iterations for the KITTI $\rightarrow$ Cityscapes experiment. Figure \ref{fig:3DCL} shows the losses of the three domain classifiers, corresponding to features F1, F2, and F3 of the baseline architecture over the first 2500 iterations of training. We can see that the losses are dissimilar after 1K iterations. This implies inconsistency among the classifiers' performance, which leads to a drop in performance. This motivated our objective in employing a unified domain classifier for all three scales. In turn, this led to the UC architecture for improving the detection performance when applied to the target domain data as shown in Tables \ref{mAP_cityscapes}, \ref{mAP_waymo} and \ref{mAP_KITTI_Cityscapes}.

\begin{figure}
\begin{center}
  \includegraphics[width=1\linewidth]{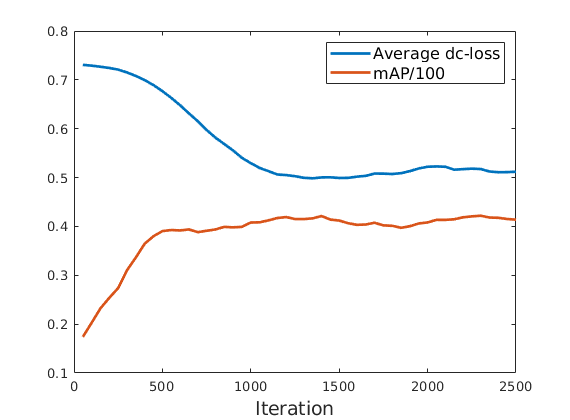}
\end{center}
  \caption{Averaged domain classifier (DC) loss over the losses of the three domain classifiers of the baseline architecture, and the detection performance in term of normalized mAP over the first 2500 iterations of training.} 
\label{fig:loss_mAP}
\end{figure}

Moreover, to study the relationship between the domain classification losses of DAN and the detection performance, we conducted a time analysis during training. We plot in Figure \ref{fig:loss_mAP} the averaged domain classifier (DC) loss over the losses of the three domain classifiers of the baseline architecture, and the detection performance in term of mAP for the KITTI $\rightarrow$ Cityscapes experiment. We normalize mAP by 100 to plot it at the same scale with average domain classifier loss. At the beginning of training, we found average DC loss starts at a higher value of 0.736. Then during training, DAN is optimized to minimize the loss while the backbone of YOLO is optimized to maximize the loss. In other words, DAN and the backbone of YOLO compete against each other. After the loss is settled around 0.5, the detection performance starts to improve because the backbone begins to produce domain invariant features at this point due to the adversarial training strategy.

\section{Conclusion}
In this paper, we proposed a multiscale domain adaptation framework for the popular state-of-the-art real time object detector YOLO. Specifically, under our MS-DAYOLO architecture, we applied domain adaptation to three different scale features within the YOLO feature extractor that are fed to the next stage. In addition to the baseline architecture of a multiscale domain adaptive network, we developed three various deep learning architectures to produce more robust domain invariant features that reduce the impact of domain shift. The proposed architectures include progressive feature reduction (PFR), unified domain classifier (UC), and the integrated architecture that combines the benefits of progressive-feature reduction and unified classifier strategies for improving the overall detection performance under the target domain. Based on various experimental results, our proposed MS-DAYOLO framework can successfully adapt YOLO to target domains without annotation. Furthermore, the proposed MS-DAYOLO architectures outperformed state-of-the-art YOLOv4 and other exciting approaches that are based on Faster R-CNN object detector under diverse testing scenarios for autonomous driving applications.


\bibliographystyle{IEEEtran}
\bibliography{ref}


\end{document}